# TSK Fuzzy System Towards Few Labeled Incomplete Multi-View Data Classification

Wei Zhang, Zhaohong Deng, *Senior Member, IEEE*, Qiongdan Lou, Te Zhang, Kup-Sze Choi, Shitong Wang

*Abstract*—TSK fuzzy system is a representative method in fuzzy systems and has received more and more attention due to its simplicity, flexibility and efficiency. With the development of data collection technology, multi-view data is becoming more and more common. To address the multi-view data, some multi-view TSK fuzzy systems have been proposed in recent years. However, most of the existing multi-view TSK fuzzy systems must rely on the larger labeled and complete multi-view data, which is a demanding condition in real-world applications. Although a few multi-view TSK fuzzy systems modeling methods have addressed issues about few labeled multi-view data or incomplete multi-view data, they cannot handle the two challenges simultaneously and still face the following challenges: first, the existing incomplete multi-view fuzzy system modeling methods usually separate the missing view imputation from model construction, which makes two processes lack effective cooperation. Second, the robustness of these methods in the few labeled data scene needs to be enhanced with more effective mechanisms. To address these issues, a novel transductive semi-supervised incomplete multi-view TSK fuzzy system modeling method (SSIMV_TSK) is proposed in this paper. First, in order to handle challenges brought by the few labeled data and missing views simultaneously, the proposed method integrates missing view imputation, pseudo label learning of unlabeled data, and fuzzy system modeling into a single process to yield a model. Then, by exploring information within and between views, two new mechanisms, i.e. the bidirectional structural preservation of instance and label, as well as the adaptive multiple alignment collaborative learning, are proposed to further improve the robustness of the model. Experimental results on real-world datasets show that the proposed method significantly outperforms state-of-the-art methods.

*Index Term*s – incomplete multi-view data, few labeled data, fuzzy systems, pseudo label learning, missing view imputation.

## I. Introduction

Fuzzy systems have been receiving great attention due to their strong data mining capability and high interpretability [1-3]. Currently, the mainstream fuzzy systems include Mamdani fuzzy system [4], Takagi-Sugeno-Kang (TSK) fuzzy system [5] and Generalized fuzzy system [6]. Among them, TSK fuzzy system has received great attention in recent years due to its high accuracy, simplicity and flexibility [7-9]. For example, by introducing the minimax probability decision technique, a minimax probability TSK system has been proposed [7]. To address very large dataset, a new TSK fuzzy system has been proposed by introducing the minimal-enclosing-ball approximation technique [8]. To improve the accuracy of TSK fuzzy system without losing its interpretability, a deep TSK fuzzy system has been proposed [9]. However, these fuzzy systems can only handle the single view data.

With the development of the data collection technology, data can be usually associated with many different types of features in real world and these data is called multi-view data. To overcome the challenges brought by multi-view data, some multi-view fuzzy systems have been proposed in recent years. For example, based on TSK fuzzy system, Jiang *et al*. introduced the collaborative learning and the maximum margin learning criteria, and proposed a multi-view TSK fuzzy system [10]. To address the multi-view epileptic EEG signal recognition problem, a multi-view TSK fuzzy system has been proposed [11]. By learning the hidden view and integrating it with visible views for modeling, a multi-view TSK fuzzy system with the cooperation between visible and hidden views was proposed [12]. In multi-view learning, not all views have the positive effect to the model learning. By introducing the adaptive threshold to filter the weak views, a view-reduction based multi-view TSK fuzzy system has been proposed [13].

It should be noted that these existing multi-view fuzzy systems are usually based on two assumptions. First, they all

This work was supported in part by the NSFC under Grant 61772239, the Six Talent Peaks Project in Jiangsu Province under Grant XYDXX-056, the Hong Kong Research Grants Council (PolyU 152006/19E), the Hong Kong Innovation and Technology Fund (MRP/015/18) and the Shanghai Municipal Science and Technology Major Project (2018SHZDZX01). (Corresponding author: Zhaohong Deng).

W. Zhang, Q. Lou, S. T. Wang are with the School of Artificial Intelligence and Computer Science, Jiangnan University and Jiangsu Key Laboratory of Media Design and Software Technology, Wuxi 214122, China (e-mail: 7201607004@stu.jiangnan.edu.cn; 6171610005@stu.jiangnan.edu.cn; wxwangst@aliyun.com).

Z. Deng is with the School of Artificial Intelligence and Computer Science, Jiangnan University, Wuxi 214122, China, and Key Laboratory of Computational Neuroscience and Brain-Inspired Intelligence (LCNBI) and ZJLab, Shanghai 200433, China. (e-mail: dengzhaohong@jiangnan.edu.cn).

K. S. Choi is with The Centre for Smart Health, the Hong Kong Polytechnic University, Hong Kong (e-mail: thomasks.choi@polyu.edu.hk).

Te Zhang is with the Lab for Uncertainty in Data and Decision Making (LUCID), School of Computer Science, University of Nottingham, Nottingham, NG8 1BB, UK (e-mail: te.zhang@nottingham.ac.uk).

assume that labeled training data are sufficient. But in real application, it is difficult and time-consuming to collect a large amount of labeled multi-view data. Second, they all assume that all views are complete, which may be not satisfied in many real-world scenarios. For example, in the camera network, due to some reasons, the camera may temporarily fail, or be blocked by some objects, resulting in instance loss [14]. Another example is document classification. When treating different languages as different views of a document, some views may be missing due to the lack of some translated versions. In these scenarios, the traditional multi-view fuzzy systems are unusable or inefficient. How to effectively model incomplete and few labeled multi-view data has become a realistic and important topic.

To address the few labeled multi-view data modeling, some semi-supervised multi-view data modeling methods have been proposed, including transductive methods and inductive methods. Transductive multi-view learning usually constructs connection between labeled data and unlabeled data to model [15, 16]. Inductive semi-supervised multi-view learning usually mines unlabeled data information to assist the classifier to model [17, 18]. In [19], a transductive multi-view TSK fuzzy system modeling method is proposed, which learns both the model and the pseudo label simultaneously, and introduces the matrix factorization to further improve the learning abilities of the model.

To address the incomplete multi-view data, some incomplete multi-view learning methods have been proposed. Missing view imputation [20-22] and common representation learning [23-25] are two types of main methods. In [26], an incomplete multi-view TSK fuzzy system modeling method is proposed. This method integrates the common representation learning and missing view imputation into one process and then combines common representation and imputed multi-view data to construct an incomplete multi-view model.

Although the above methods can address the challenge brought by few labeled multi-view data and incomplete multi-view data to a certain extent, they still have the following problems: (1) They cannot address incomplete and few labeled multi-view data simultaneously, but the incomplete and few labeled multi-view data is widely existing in real application scenarios. For example, in web classification, web content and title can be regarded as two views. It is difficult to label all web pages and the views may be missing due to collection failure. (2) In the above incomplete multi-view learning methods, model construction and missing view imputation are usually separated, which prevent two processes from negotiating with each other to achieve more efficient learning. (3) The robustness of these existing methods still needs to improve, especially for the scenes of few labeled data.

In this paper, a novel transductive semi-supervised incomplete multi-view TSK fuzzy system modeling method (SSIMV_TSK) is proposed to address the above issues. First, a new transductive semi-supervised incomplete multi-view model with good interpretability is constructed based on TSK fuzzy system. In this model, pseudo label learning, missing view imputation, and model learning are integrated into one optimization framework, which not only reduces the dependency on labeled data, but also make the model construction and missing view imputation benefit from each other. Second, to improve the quality of imputed views and the accuracy of the pseudo labels, two new mechanisms, i.e., the bidirectional structural preservation of instances and labels, and the adaptive multiple alignment collaborative learning, are proposed. These mechanisms can effectively mine the correlation information within views, as well as the consistency and complementarity information between views. Third, to obtain the optimal view weight allocation, the adaptive view weighting mechanism is introduced into the model, which also further improves the adaptability of the model.

The main contributions of this paper are as follows:

1) A novel transductive semi-supervised incomplete multi-view TSK fuzzy system is proposed, which unifies the missing views imputation, pseudo label learning, and model learning into one optimization framework. The new framework makes each learning task benefit from the others.

2) The bidirectional structural preservation of instances and labels, as well as adaptive multiple alignment collaborative learning are proposed to improve the quality of imputed views and pseudo labels. Therefore, the multi-view learning can be enhanced accordingly.

3) Extensive experiments on benchmark datasets were carried out and the results demonstrated that our method outperforms several baseline methods and state-of-the-art methods.

The rest of this paper is organized as follows. In Section II, the related work is reviewed. Section III provides the details of the semi-supervised incomplete multi-view fuzzy system. In Section IV, the proposed method is extensively evaluated on real-world datasets. Conclusion is given in Section V.

## II. RELATED WORK

### A. Takagi-Sugeno-Kang Fuzzy System

Given a TSK fuzzy system, the *k*-th fuzzy inference rule is often defined as follows:

$$R^k: IF\ x_1\ is\ A_1^k\ \wedge \cdots \wedge\ x_d\ is\ A_d^k$$
$$THEN\ f_k^1(x) = p_0^{k,1} + p_1^{k,1}x_1 + \cdots + p_d^{k,1}x_d,$$
$$\vdots$$
$$f_k^c(x) = p_0^{k,c} + p_1^{k,c}x_1 + \cdots + p_d^{k,c}x_d,$$
$$\vdots$$
$$f_k^C(x) = p_0^{k,C} + p_1^{k,C}x_1 + \cdots + p_d^{k,C}x_d,$$

$$k = 1,2,\ldots,K, c = 1,2,\ldots,C \quad (1)$$

where $\mathbf{x} = [x_1, x_2, \ldots, x_d]$ is the input vector, $A_d^k$ is a fuzzy subset of the input vector, $\wedge$ is a fuzzy conjunction operator, $K$ is the number of fuzzy rules, and $C$ is the number of outputs. When the multiplicative conjunction, multiplicative implication, and additive combination are adopted, the outputs of a classical TSK fuzzy system is given as follows:

$$\mathbf{y}^o = [y^{o,1}, \ldots, y^{o,C}] = \sum_{k=1}^{K} \frac{\mu^k(\mathbf{x})\mathbf{f}_k(\mathbf{x})}{\sum_{k'}^{K} \mu^{k'}(\mathbf{x})} = \sum_{k=1}^{K} \tilde{\mu}^k(\mathbf{x})\mathbf{f}_k(\mathbf{x}) \quad (2a)$$

$$\mathbf{f}_k(\mathbf{x}) = [f_k^1(\mathbf{x}), f_k^2(\mathbf{x}), \ldots, f_k^C(\mathbf{x})] \quad (2b)$$

where $\mu^k(\mathbf{x})$ and $\tilde{\mu}^k(\mathbf{x})$ are the firing strength and the normalized firing strength of a rule, and they can be calculated as follows:

$$\mu^k(\mathbf{x}) = \prod_{i=1}^{d} \mu_{A_i^k}(x_i) \quad (2c)$$

$$\tilde{\mu}^k(\mathbf{x}) = \frac{\mu^k(\mathbf{x})}{\sum_{k'=1}^{K} \mu^{k'}(\mathbf{x})} \quad (2d)$$

where $\mu_{A_i^k}(x_i)$ in (2c) is the membership of the corresponding fuzzy set $A_i^k$. The membership can be computed as follows if the Gaussian membership function is adopted:

$$\mu_{A_i^k}(x_i) = exp\left(\frac{-(x_i - e_i^k)^2}{2\delta_i^k}\right) \quad (2e)$$

where parameters $e_i^k$ and $\delta_i^k$ are the center and width of the Gaussian function, respectively. In classical TSK fuzzy systems, the two parameters are also called the antecedent parameters. They can be estimated with different strategies, typically using clustering techniques, such as Fuzzy C-Means Clustering (FCM) [27]. However, since random initialization is involved in FCM, the results may be unstable. To overcome this shortcoming, more stable clustering algorithms such as Var-Part [28] can be adopted.

Once the antecedent parameters are estimated, the outputs of the TSK fuzzy system in (2a) can be expressed as follows:

$$\mathbf{y}^o = \mathbf{x}_g \mathbf{P}_g \in R^{1 \times C} \quad (3a)$$

where $\mathbf{x}_g$ and $\mathbf{P}_g$ are defined as follows:

$$\mathbf{x}_e = [1, \mathbf{x}] \quad (3b)$$
$$\tilde{\mathbf{x}}^k = \tilde{\mu}^k(\mathbf{x})\mathbf{x}_e \quad (3c)$$
$$\mathbf{x}_g = ((\tilde{\mathbf{x}}^1), (\tilde{\mathbf{x}}^2), \ldots, (\tilde{\mathbf{x}}^K)) \quad (3d)$$
$$\mathbf{p}_k^c = (p_0^{k,C}, p_1^{k,C}, \ldots, p_d^{k,C}) \quad (3e)$$
$$\mathbf{P}^c = (\mathbf{p}_1^c, \mathbf{p}_2^c, \ldots, \mathbf{p}_K^c) \quad (3f)$$
$$\mathbf{P}_g = ((\mathbf{P}^1)^T, (\mathbf{P}^2)^T, \ldots, (\mathbf{P}^C)^T) \quad (3g)$$

### B. Multi-view Takagi-Sugeno-Kang Fuzzy System

Given a multi-view dataset with $N$ instances and $V$ views $\{\mathbf{X}^v = [\mathbf{x}_1; \mathbf{x}_2; \ldots; \mathbf{x}_N] \in R^{N \times d^v}, v = 1,2,\ldots,V\}$, where $d^v$ is the dimension number of the $v$-th view, multi-view TSK fuzzy system usually contains the following optimization objectives [11]:

$$\min_{\mathbf{P}_g^v, a^v} \sum_{v=1}^{V} \sum_{i=1}^{N} a^v \|\mathbf{x}_{g,i}^v \mathbf{P}_g^v - \mathbf{y}_i\|^2 + \beta_1 \sum_{v=1}^{V} \|\mathbf{P}_g^v\|^2 +$$
$$\beta_3 \sum_{v=1}^{V} \sum_{i=1}^{N} \left\|\mathbf{x}_{g,i}^v \mathbf{P}_g^v - \frac{1}{V-1}\sum_{t=1, t \neq v}^{V} \mathbf{x}_{g,i}^t \mathbf{P}_g^t\right\|^2 +$$
$$\beta_2 \sum_{v=1}^{V} a^v \ln a^v$$
$$s.t. \ a^v \geq 0, \ \sum_{v=1}^{V} a^v = 1 \quad (4)$$

where $\mathbf{x}_{g,i}^v \in R^{1 \times d_g^v}$ is the mapping of the original data $\mathbf{x}_i^v \in R^{1 \times d^v}$ in the new feature space through the fuzzy rules using (3b)-(3d), and $\mathbf{x}_i^v$ is the $i$-th instance of $\mathbf{X}^v$. $d_g^v$ is the number of features in the new space. $\mathbf{P}_g^v \in R^{d_g^v \times C}$ is the matrix expression of the consequent parameters of the $v$-th view. $C$ is the number of classes. $\mathbf{y}_i \in R^{1 \times C}$ is the label vector of the $i$-th instance in the multi-view data. For example, $\mathbf{y}_i = [0, 1, 0]$ denotes that $\mathbf{x}_{g,i}^v$ belongs to the second class. $a^v$ is the view weight. The first term $\|\mathbf{x}_{g,i}^v \mathbf{P}_g^v - \mathbf{y}_i\|^2$ is the training error across different views. The second term $\|\mathbf{P}_g^v\|^2$ is the regularization term, and it is used to improve the robustness of the model. The third term $\left\|\mathbf{x}_{g,i}^v \mathbf{P}_g^v - \frac{1}{V-1}\sum_{t=1, t \neq v}^{V} \mathbf{x}_{g,i}^t \mathbf{P}_g^t\right\|^2$ is the cooperation term, and it is used to ensure consistency between the outputs of the multiple views and to mine consistent information between views. The fourth term $\sum_{v=1}^{V} a^v \ln a^v$ is the negative Shannon entropy, which is introduced to adjust the weights of views adaptively. $\beta_1$, $\beta_2$ and $\beta_3$ are the regularization parameters.

Obviously, as shown in (4), the method proposed in [26] cannot utilize the unlabeled data and explore the incomplete multi-view data effectively. Two improved versions have been proposed in [26] and [19]. [26] can handle the incomplete multi-view problem satisfactorily, but it separates the missing view imputation and model learning, and thus the effectiveness needs to be improved. [19] can reduce the dependency on labeled data by introducing transductive learning, but it is unable to directly explore the incomplete multi-view data. Therefore, when we face the incomplete, few labeled multi-view data, a new model is needed. In this paper, we unify the missing view imputation, pseudo label learning, and model constructing into one framework, which enables the model to directly address these three tasks simultaneously and then make each task benefit from the others. Meanwhile, by mining the between-views and within-views information, the bidirectional structural preservation of instances and labels, as well as the adaptive multiple alignment collaborative learning are proposed to improve the quality of imputed views and the accuracy of pseudo labels.

### C. Multi-view Semi-supervised Learning

To address the few labeled multi-view learning problem, many methods have been proposed in recent years and the existing methods can be divided into the following two categories transductive method and inductive methods.

Transductive semi-supervised multi-view learning: this category of methods usually construct the connection between labeled data and unlabeled. For example, Karasuyama *et al*. integrated multiple adjacency graphs for label propagation and proposed Sparse Multiple Graph Integration (SMGI) [15]. By optimizing the adjacency graph and label simultaneously, Nie *et al*. proposed Multi-View Learning with Adaptive Neighbors (MLANs) [16] and introduced an adaptive weighting mechanism to improve the performance of the model.

Inductive semi-supervised multi-view learning: this category of methods mainly mine unlabeled data information to assist the classifier learning. For example, Tao *et al*. proposed a multi-view semi-supervised method (MVAR) [17] by constructing a global regression model for each view and getting the final decision by weighted combining the decision of each view. Traditional semi-supervised support vector machines can only handle two-view datasets [29, 30], and have high computational complexity. To address these issues, Sun *et al*. proposed a general Multi-View Semi-Supervised Support Vector Machines (GMvLapSVM) [18].

Although these two categories of multi-view semi-supervised learning methods can handle the few labeled multi-view data to a certain extent, they cannot handle the incomplete multi-view data and most of them neglect the interpretability of model. In this study, the semi-supervised learning with the interpretable model for incomplete multi-view data will be studied.

### D. Incomplete Multi-view Learning

Missing view imputation and common representation learning are two main strategies to address the incomplete multi-view learning problem.

Missing view imputation-based methods usually mine the information within views or between views for imputing. For example, by only exploring the information within views, Talyanskaya et al. used the *k*-nearest neighbor (KNN) to mine the *k*-th similar instances to impute missing instances [20]. Bhadra *et al*. mined the correlation information within and between views in the kernel space to impute missing views [21]. Based on the identically distributed constraint between subspace and views, Zhang *et al*. used complete subspace to impute missing views [22]. Common representation learning-based methods usually learn the representation between views. For example, By exploring similar information within views, Wen *et al*. jointed the nonnegative matrix factorization (NMF) and missing view imputation into one optimization process [23]. To further improve the quality of the learned common representation, some nonlinear data mining techniques were used in common representation learning. For example, Liu *et al.* used multiple kernel technology to integrate the missing view imputation and common representation learning in one process [24] . By combining the semi-supervised deep matrix decomposition and correlation subspace learning, Xue *et al*. proposed a novel incomplete multi-view learning method [25].

These two categories of methods are two-step methods, i.e., the missing view imputation or common representation learning is separated from the subsequent model construction, which makes the missing view imputation or common representation learning lack pertinence for the model construction of the given modeling task. Unlike the above existing methods, a new one-step method will be proposed in this study.

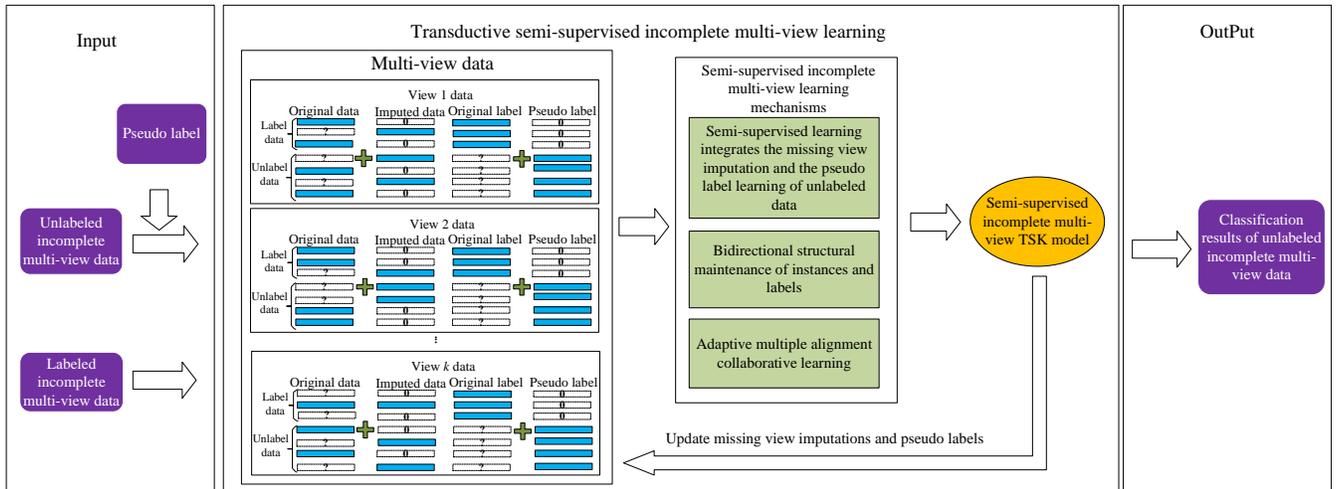

Fig. 1. The framework of the proposed semi-supervised incomplete multi-view learning for TSK fuzzy systems.

## III. SEMI-SUPERVISED INCOMPLETE MULTI-VIEW FUZZY SYSTEM

### A. The Framework of Semi-Supervised Incomplete Multi-View Fuzzy System

As discussed in Introduction, the following three issues in multi-view fuzzy system modeling should be further studied: 1) the existing methods cannot address incomplete and few labeled multi-view data simultaneously; 2) the existing incomplete multi-view TSK fuzzy system separates the model construction from missing view imputation, which prevents the two processes negotiating with each other; 3) the robustness of these methods still needs to improve.

To address the above challenges, a novel transductive semi-supervised incomplete multi-view TSK fuzzy system modeling method, i.e., SSIMV_TSK is proposed. The framework of SSIMV_TSK is shown in Fig.1. As shown in Fig. 1, the process of SSIMV_TSK is as follows: First, the randomly generated pseudo labels are assigned to the unlabeled data, which are integrated with the labeled data to form the initial incomplete multi-view data for learning. Second, in semi-supervised incomplete multi-view learning stage, missing view imputation, pseudo label learning and model building are conducted simultaneously as a whole to make each of them benefit from the others. Meanwhile, the bidirectional structural preservation of instances and labels, as well as the adaptive multiple alignment collaborative learning are proposed to improve the quality of imputed views and pseudo labels. Finally, the classification results of unlabeled data are output.

### B. Semi-Supervised Incomplete Multi-View Fuzzy System

To address the few labeled multi-view data learning problem, we improve (4) and propose the following objective function:

$$\min_{\mathbf{P}_g^v, \mathbf{y}_i^u, a^v} \sum_{v=1}^V \sum_{i=1}^N a^v \left\| \tilde{\mathbf{x}}_{g,i}^v \mathbf{P}_g^v - \tilde{\mathbf{y}}_i \right\|^2 + \beta_1 \sum_{v=1}^V \left\| \mathbf{P}_g^v \right\|^2 + \beta_2 \sum_{v=1}^V a^v \ln a^v$$

$$s.t. \, a^v \geq 0, \, \sum_{v=1}^V a^v = 1, \mathbf{y}_i^u \geq 0, \, \sum_i^c \mathbf{y}_i^u = 1 \quad (5)$$

where $\tilde{\mathbf{x}}_{g,i}^v \in R^{1 \times d_g^v}$ is the $i$-th instance of $\tilde{\mathbf{X}}_g^v = [\mathbf{X}_g^{l,v}; \mathbf{X}_g^{u,v}] \in R^{N \times d_g^v}$, $\mathbf{X}_g^{l,v} = [\mathbf{x}_{g,1}^{l,v}; \mathbf{x}_{g,2}^{l,v}; \ldots; \mathbf{x}_{g,N_l}^{l,v}] \in R^{N_l \times d_g^v}$ is the labeled data of the $v$-th view, $\mathbf{X}_g^{u,v} = [\mathbf{x}_{g,1}^{u,v}; \mathbf{x}_{g,2}^{u,v}; \ldots; \mathbf{x}_{g,N_u}^{u,v}] \in R^{N_u \times d_g^v}$ is the unlabeled data of the $v$-th view, and $N=N_l+N_u$ is the total number of instances with $N_l$ and $N_u$ as the numbers of the labeled and unlabeled data, respectively. $\tilde{\mathbf{y}}_i \in R^{1 \times C}$ is the $i$-th label vector in $\tilde{\mathbf{Y}} = [\mathbf{Y}^l; \mathbf{Y}^u] = [\tilde{\mathbf{y}}_1; \tilde{\mathbf{y}}_2; \ldots; \tilde{\mathbf{y}}_N] \in R^{N \times C}$. $\mathbf{Y}^l = [\mathbf{y}_1^l; \mathbf{y}_2^l; \ldots; \mathbf{y}_{N_l}^l] \in R^{N_l \times C}$ is the label matrix of the labeled data. $\mathbf{Y}^u = [\mathbf{y}_1^u; \mathbf{y}_2^u; \ldots; \mathbf{y}_{N_u}^u] \in R^{N_u \times C}$ is the pseudo label matrix of the unlabeled data. $\mathbf{P}_g^v \in R^{d_g^v \times C}$ is the consequent parameter matrix of the $v$-th view. Compared with (4), the pseudo labels learning and unlabeled data are integrated with labeled data for model training in (5), which greatly improves the learning ability on few labeled multi-view data. Besides, (5) is further improved as follows to address the incomplete multi-view problem:

$$\min_{\mathbf{P}_g^v, \tilde{\mathbf{h}}_i^v, \mathbf{y}_i^u, a^v} \sum_{v=1}^V \sum_{i=1}^N a^v \left\| \widetilde{w}_{i,i}^v \left( (\tilde{\mathbf{x}}_{g,i}^v + \tilde{e}_{i,i}^v \tilde{\mathbf{h}}_i^v) \mathbf{P}_g^v - \tilde{\mathbf{y}}_i \right) \right\|^2 + \beta_1 \sum_{v=1}^V \left\| \mathbf{P}_g^v \right\|^2 + \beta_2 \sum_{v=1}^V a^v \ln a^v$$

$$s.t. \, a^v \geq 0, \, \sum_{v=1}^V a^v = 1, \mathbf{y}_i^u \geq 0, \, \sum_i^c \mathbf{y}_i^u = 1 \quad (6)$$

where $\tilde{\mathbf{x}}_{g,i}^v + \tilde{e}_{i,i}^v \tilde{\mathbf{h}}_i^v$ is imputed instance in the fuzzy feature space. $\tilde{e}_{i,i}^v$ is the $i$-th element on the diagonal of $\tilde{\mathbf{E}}^v = [[\mathbf{E}^{l,v}, \mathbf{M}^{l,v}]; [\mathbf{M}^{u,v}, \mathbf{E}^{u,v}]] \in \mathbf{R}^{N \times N}$. $\mathbf{M}^{l,v} \in \mathbf{R}^{N_l \times N_u}$, $\mathbf{M}^{u,v} \in \mathbf{R}^{N_u \times N_l}$ are the zero matrix. $\tilde{\mathbf{E}}^v$ is a diagonal matrix of the $v$-th view and represents the indicator matrix. $\mathbf{E}^{l,v} \in R^{N_l \times N_l}$ and $\mathbf{E}^{u,v} \in R^{N_u \times N_u}$ are indicator matrix of the labeled data and unlabeled data of the $v$-th view, respectively. $\tilde{e}_{i,i}^v$ can be defined as follow:

$$\tilde{e}_{i,i}^v = \begin{cases} 1, & \text{if } i\text{-th instance is the } i\text{-th missing} \\ & \text{instance in the } v\text{-th view} \\ 0, & \text{otherwise} \end{cases} \quad (7)$$

$\tilde{\mathbf{h}}_i^v \in R^{1 \times d_g^v}$ is the $i$-th instance of $\tilde{\mathbf{H}}^v$. $\tilde{\mathbf{H}}^v = [\mathbf{H}^{l,v}; \mathbf{H}^{u,v}] \in \mathbf{R}^{N \times d_g^v}$ is the error matrix of the $v$-th view, and it is used to impute the missing values. $\mathbf{H}^{l,v} = [\mathbf{h}_1^{l,v}; \mathbf{h}_2^{l,v}; \ldots; \mathbf{h}_{N_l}^{l,v}] \in R^{N_l \times d_g^v}$ and $\mathbf{H}^{u,v} = [\mathbf{h}_1^{u,v}; \mathbf{h}_2^{u,v}; \ldots; \mathbf{h}_{N_u}^{u,v}] \in R^{N_u \times d_g^v}$ are the error matrix of the labeled data and unlabeled data of the $v$-th view, respectively. $\widetilde{w}_{i,i}^v$ is the $i$-th instance on the diagonal of $\widetilde{\mathbf{W}}^v$. $\widetilde{\mathbf{W}}^v = [[\mathbf{W}^{l,v}, \mathbf{M}^{l,v}]; [\mathbf{M}^{u,v}, \mathbf{W}^{u,v}]] \in \mathbf{R}^{N \times N}$ is the sample weight matrix of each view. $\mathbf{W}^{l,v} \in R^{N_l \times N_l}$ and $\mathbf{W}^{u,v} \in R^{N_u \times N_u}$ are the sample weight matrix of the labeled data and unlabeled data of the $v$-th view. $\widetilde{w}_{i,i}^v$ is given by:

$$\widetilde{w}_{i,i}^v = \begin{cases} 1, & \text{if the } v\text{-th view contains the } i\text{-th instance} \\ w^v, & \text{otherwise} \end{cases} \quad (8)$$

where $w^v$ is the weight of the imputed view, defined as the ratio of the number of available instances to the total number of instances. Through (7), we integrate the missing view imputation, pseudo label learning, and model building into one process, which makes each task benefit from the others in the learning procedure.

### C. Bidirectional Structural Preservation of Instances and Labels

To alleviate the incomplete and few labeled multi-view data learning problem, two issues need to be solved. The first issue is how to mine the similarity information to alleviate the negative impact of the missing view. Existing methods, such as [20, 23], mine the similarity between complete instances and missing instances to guide missing instances imputation. However, they ignore the connection between labels and instances, which can easily lead imputed instances to deviate from their original category. Further, as the imputed instances deviate from their original category, the discriminability information and the consistency information between views will greatly decline. The second issue is how to effectively learn pseudo labels to alleviate the dependency on a larger number of labeled data. To address this problem, some methods introduce the matrix decomposition to transform the hard pseudo label matrix into soft pseudo label matrix [31-33], but the internal information of unlabeled multi-view data is ignored. Besides, these methods need to introduce additional decomposition parameter.

To overcome the above issues, a new mechanism, called the bidirectional structural preservation of instances and labels is proposed, which can be formulated as follows:

$$\min_{\mathbf{y}_i^u, \tilde{\mathbf{h}}_i^v, \mathbf{P}_g^v} \sum_{v=1}^{V} \sum_{i=1}^{N_u} \sum_{j=1}^{N_u} \|\mathbf{y}_i^u - \mathbf{y}_j^u\| s_{i,j}^{u,v} + \sum_{v=1}^{V} \sum_{i=1}^{N} \sum_{j=1}^{N} \|(\tilde{\mathbf{x}}_{g,i}^v + \tilde{e}_{i,i}^v \tilde{\mathbf{h}}_i^v) \mathbf{P}_g^v - (\tilde{\mathbf{x}}_{g,j}^v + \tilde{e}_{j,j}^v \tilde{\mathbf{h}}_j^v) \mathbf{P}_g^v\| z_{i,j}$$
$$s.t. \ \mathbf{y}_i^u \geq 0, \ \sum_{i}^{c} \mathbf{y}_i^u = 1 \quad (9)$$

Based on (9) the structural information of instances and labels can be preserved with a bidirectional way, which is further explained as follows:

1) The first term in (9) is the structural preservation of pseudo labels in the $v$-th view using the similarity information of instances. $\mathbf{y}_i^u$ is the pseudo label vetor of the $i$-th instance in unlabeled dataset. $\mathbf{S}^{u,v} \in R^{N_u \times N_u}$ is the similarity matrix unlabeled data in the $v$-th view, and $s_{i,j}^{u,v}$ is described as follows:

$$s_{i,j}^{u,v} = \begin{cases} \kappa(\mathbf{q}_i^{u,v}, \mathbf{q}_j^{u,v}), & \text{if } i\text{-th instance is the } k\text{-nearest} \\ & \text{neighbor of } j\text{-th instance and the } i\text{-th or} \\ & j\text{-th instance is complete instance} \\ 0, & \text{otherwise} \end{cases} \quad (10)$$

where $\mathbf{q}_i^{u,v} = \mathbf{x}_{g,i}^{u,v} + e_{i,i}^{u,v} \mathbf{h}_i^{u,v}$ is the imputed instance, and $\kappa(*,*)$ is the kernel function.

2) The second term in (9) is the structural preservation of projected instances using the similarity information of labels. It uses the similar information of the label to impute the missing view and constrain the consistency of the consequent parameters. $(\tilde{\mathbf{x}}_{g,i}^v + \tilde{e}_{i,i}^v \tilde{\mathbf{h}}_i^v) \mathbf{P}_g^v$ is the predicted label vector of the $i$-th instance, which can be viewed as the projected vector of the imputed instance $(\tilde{\mathbf{x}}_{g,i}^v + \tilde{e}_{i,i}^v \tilde{\mathbf{h}}_i^v)$. $\mathbf{Z} \in R^{N \times N}$ is the similarity matrix of labels in the $v$-th view, and $z_{i,j}$ is defined as follows:

$$z_{i,j} = \begin{cases} 1, & \text{if } i\text{-th label and } j\text{-th label vector are true labels and} \\ & \text{have same label values} \\ \kappa(\tilde{\mathbf{y}}_i, \tilde{\mathbf{y}}_j), & \text{if } i\text{-th label vector is the } k\text{-nearest neighbor} \\ & \text{of } j\text{-th label vector and } i\text{-th and } j\text{-th label vectors} \\ & \text{are not all true} \\ 0, & \text{otherwise} \end{cases} \quad (11)$$

where $\kappa(*,*)$ is the kernel function.

The objective of (9) has realized the bidirectional structural preservation of instance and label, which makes the pseudo label learning and the missing view imputation mutually improved simultaneously.

### D. Adaptive Multiple Alignment Collaborative Learning

Collaborative learning mechanism is usually introduced to mine the consistency and complementary information of the multi-view data. A commonly used mechanism can be formulated as follows [11]:

$$\min_{\mathbf{P}_g^v} \sum_{v=1}^{V} \sum_{i=1}^{N} \left\| \mathbf{x}_{g,i}^v \mathbf{P}_g^v - \frac{1}{V-1} \sum_{t=1, t \neq v}^{V} \mathbf{x}_{g,i}^t \mathbf{P}_g^t \right\| \quad (12)$$

where $\mathbf{x}_{g,i}^v$ is the $i$-th instance of the $v$-th view. $\mathbf{P}_g^v$ is the consequent parameters of model for the $v$-th view. In this strategy, the prediction results of each view are aligned with the other views. However, this method has two drawbacks. First, the discriminability of each view is different, and thus it is not an optimal choice to align each view with the other views equally. Especially, this drawback is more obvious when facing incomplete multi-view data. For example, when the missing instances in one view are more than those in the other views, the discriminability of this view is obviously less than the others. In this case, if we treat all views equally, the result of collaborative learning will unsatisfactory. Second, traditional collaborative learning methods only align the prediction results of the same instance in different views, and cannot mine the deep consistent information between different instances, i.e., the consistency information between the instance and the several similar instances in other views, which loses lots of discriminant information. To address the above issues, we proposed a new adaptive multiple alignment collaborative learning strategy as follows:

$$\min_{\tilde{\mathbf{h}}_i^v, \mathbf{P}_g^v} \sum_{v=1}^{V} \sum_{i=1}^{N} \sum_{j=1}^{N} s_{i,j}^v \left\| (\tilde{\mathbf{x}}_{g,i}^v + \tilde{e}_{i,i}^v \tilde{\mathbf{h}}_i^v) \mathbf{P}_g^v - \sum_{t=1, t \neq v}^{V} a^t (\tilde{\mathbf{x}}_{g,j}^t + \tilde{e}_{j,j}^t \tilde{\mathbf{h}}_j^t) \mathbf{P}_g^t \right\| \quad (13)$$

where $s_{i,j}^v$ is the element at the $i$-th row and $j$-th column of $\mathbf{S}^v \in R^{N \times N}$. $\mathbf{S}^v$ is defined as in (10), and the only difference is that it is constructed on all data. $a^t$ is the view weight of $t$-th view. By introducing the similarity matrix, (13) mines the consistency information between the instance and the several similar instances in other views, and realizes the deep correlation information mining. Meanwhile, (13) introduces the view weight and improves the adaptability of the mined information. Besides, in incomplete multi-view scenes, though mining the consistency and complementarity information, (13) can improve the quality of consequent parameters and the imputed missing views simultaneously [21].

### E. Overall Objective Function

Based on the above analysis, the objective function of the proposed SSIMV_TSK can be defined as follows:

$$\min_{\mathbf{P}_g^v, \tilde{\mathbf{h}}_i^v, \mathbf{y}_i^u, a^v} J_{IMVSS\_TSK}(\mathbf{P}_g^v, \tilde{\mathbf{h}}_i^v, \mathbf{y}_i^u, a^v) = \Gamma + \Delta + \Theta$$

$$s.t. a^v \geq 0, \sum_{v=1}^{V} a^v = 1, y_{i,j}^u \geq 0, \sum_{j=1}^{c} y_{i,j}^u = 1 \quad (14a)$$

$$\Gamma = \sum_{v=1}^{V} \sum_{i=1}^{N} a^v \left\| \widetilde{w}_{i,i}^v \left( (\tilde{\mathbf{x}}_{g,i}^v + \tilde{e}_{i,i}^v \tilde{\mathbf{h}}_i^v) \mathbf{P}_g^v - \tilde{\mathbf{y}}_i \right) \right\|^2 +$$
$$\beta_1 \sum_{v=1}^{V} \left\| \mathbf{P}_g^v \right\|^2 + \beta_2 \sum_{v=1}^{V} a^v \ln a^v \quad (14b)$$

$$\Delta = \beta_4 \sum_{v=1}^{V} \sum_{i=1}^{N_u} \sum_{j=1}^{N_u} s_{i,j}^{u,v} \left\| \mathbf{y}_i^u - \mathbf{y}_j^u \right\| +$$
$$\beta_3 \sum_{v=1}^{V} \sum_{i=1}^{N} \sum_{j=1}^{N} z_{i,j} \left\| (\tilde{\mathbf{x}}_{g,i}^v + \tilde{e}_{i,i}^v \tilde{\mathbf{h}}_i^v) \mathbf{P}_g^v - (\tilde{\mathbf{x}}_{g,j}^v + \tilde{e}_{j,j}^v \tilde{\mathbf{h}}_j^v) \mathbf{P}_g^v \right\| \quad (14c)$$

$$\Theta = \beta_5 \sum_{v=1}^{V} \sum_{i=1}^{N} \sum_{j=1}^{N} s_{i,j}^v \left\| (\tilde{\mathbf{x}}_{g,i}^v + \tilde{e}_{i,i}^v \tilde{\mathbf{h}}_i^v) \mathbf{P}_g^v - \sum_{t=1, t \neq v}^{V} a^t (\tilde{\mathbf{x}}_{g,j}^t + \tilde{e}_{j,j}^t \tilde{\mathbf{h}}_j^t) \mathbf{P}_g^t \right\| \quad (14d)$$

where $\beta_1, \beta_2, \beta_3, \beta_4, \beta_5$ are used to control the impact of the corresponding terms. These hyperparameters can be set manually or be determined by cross-validation strategy. Further explanations about the objective function are given below.

1) (14b) is the training error across different views. $\sum_{v=1}^{V} a^v \ln a^v$ is introduced to adjust weights of views adaptively. The appropriate weights of views can then be learned to improve the robustness of the model.

2) The first term of (14c), i.e., $\sum_{v=1}^{V} \sum_{i=1}^{N_u} \sum_{j=1}^{N_u} \left\| \mathbf{y}_i^u - \mathbf{y}_j^u \right\| s_{i,j}^{u,v}$, links the unlabeled instances and their pseudo labels, which preserves the structure of labels using the structure information of instances, and then alleviates the dependency of a large amount of labeled data.

3) By mining the correlation information among instances with the similar label within each view, the second term of (14c), i.e., $\sum_{v=1}^{V} \sum_{i=1}^{N} \sum_{j=1}^{N} \left\| (\tilde{\mathbf{x}}_{g,i}^v + \tilde{e}_{i,i}^v \tilde{\mathbf{h}}_i^v) \mathbf{P}_g^v - (\tilde{\mathbf{x}}_{g,j}^v + \tilde{e}_{j,j}^v \tilde{\mathbf{h}}_j^v) \mathbf{P}_g^v \right\| z_{i,j}$, realizes the structural preservation of projected instances using the structure information of labels, which not only ensures the consistence of the topological structure between instances and labels, but also improves the quality of imputed missing views.

4) By introducing the similarity matrix and view weight, (14d) realizes the adaptive multiple alignment collaborative learning, which not only mines consistent information between views and ensures consistency between the outputs of multiple views, but also mines the complementarity information to improve imputed views.

5) Combing the second term of (14c) and the (14d), the specific information within views and the complementarity information between views can be mined to greatly improve the quality of the imputed view. Combing (14d) and the first term of (14c), the consistency of pseudo labels and the outputs of each view can be ensured, which enables the model to make full use of unlabeled data and improves the robustness of the model.

*F. Solution of Semi-Supervised Incomplete Multi-view TSK Fuzzy System*

The optimization problem in (14a) is non-convex and can be solved iteratively. The optimization procedure is described as follows.

1) Update $a^v$ with $\tilde{\mathbf{h}}_i^v$, $\mathbf{y}_i^u$ and $\mathbf{P}_g^v$ fixed

Keeping $\tilde{\mathbf{h}}_i^v$, $\mathbf{y}_i^u$ and $\mathbf{P}_g^v$ fixed, the following objective function should be minimized:

$$\min_{a^v} \sum_{v=1}^{V} \sum_{i=1}^{N} a^v \left\| \widetilde{w}_{i,i}^v \left( (\tilde{\mathbf{x}}_{g,i}^v + \tilde{e}_{i,i}^v \tilde{\mathbf{h}}_i^v) \mathbf{P}_g^v - \tilde{\mathbf{y}}_i \right) \right\|^2 +$$
$$\beta_2 \sum_{v=1}^{V} a^v \ln a^v$$
$$s.t. a^v \geq 0, \sum_{v=1}^{V} a^v = 1 \quad (15)$$

Taking the derivative of (15) with respect to $a^v$ and setting it to zero, the update rule for $a^v$ can be obtained as follows:

$$a^v = \frac{\exp\left(-\sum_{i=1}^{N} \left\| \widetilde{w}_{i,i}^v \left( (\tilde{\mathbf{x}}_{g,i}^v + \tilde{e}_{i,i}^v \tilde{\mathbf{h}}_i^v) \mathbf{P}_g^v - \tilde{\mathbf{y}}_i \right) \right\|^2 / \beta_2\right)}{\sum_{l}^{V} \exp\left(-\sum_{i=1}^{N} \left\| \widetilde{w}_{i,i}^l \left( (\tilde{\mathbf{x}}_{g,i}^l + \tilde{e}_{i,i}^l \tilde{\mathbf{h}}_i^l) \mathbf{P}_g^l - \tilde{\mathbf{y}}_i \right) \right\|^2 / \beta_2\right)} \quad (16)$$

2) Update $\mathbf{P}_g^v$ with $\tilde{\mathbf{h}}_i^v$, $\mathbf{y}_i^u$ and $a^v$ fixed

Keeping $\tilde{\mathbf{h}}_i^v$, $\mathbf{y}_i^u$ and $a^v$ fixed, the following objective function should be minimized:

$$\min_{\mathbf{P}_g^v} \sum_{v=1}^{V} \sum_{i=1}^{N} a^v \left\| \widetilde{w}_{i,i}^v \left( (\tilde{\mathbf{x}}_{g,i}^v + \tilde{e}_{i,i}^v \tilde{\mathbf{h}}_i^v) \mathbf{P}_g^v - \tilde{\mathbf{y}}_i \right) \right\|^2 +$$
$$\beta_3 \sum_{v=1}^{V} \sum_{i=1}^{N} \sum_{j=1}^{N} z_{i,j} \left\| (\tilde{\mathbf{x}}_{g,i}^v + \tilde{e}_{i,i}^v \tilde{\mathbf{h}}_i^v) \mathbf{P}_g^v - (\tilde{\mathbf{x}}_{g,j}^v + \tilde{e}_{j,j}^v \tilde{\mathbf{h}}_j^v) \mathbf{P}_g^v \right\| + \beta_5 \sum_{v=1}^{V} \sum_{i=1}^{N} \sum_{j=1}^{N} s_{i,j}^v \left\| (\tilde{\mathbf{x}}_{g,i}^v + \tilde{e}_{i,i}^v \tilde{\mathbf{h}}_i^v) \mathbf{P}_g^v - \sum_{t=1, t \neq v}^{V} a^t (\tilde{\mathbf{x}}_{g,j}^t + \tilde{e}_{j,j}^t \tilde{\mathbf{h}}_j^t) \mathbf{P}_g^t \right\| +$$
$$\beta_1 \sum_{v=1}^{V} \left\| \mathbf{P}_g^v \right\|^2 \quad (17)$$

Taking the derivative of (17) with respect to $\mathbf{P}_g^v$ and setting it to zero, the update rule for $\mathbf{P}_g^v$ can be obtained as follows:

$$\mathbf{P}_g^v = \left( \beta_1 \mathbf{I} + a^v \sum_{i=1}^{N} \mathbf{q}_i^T (\widetilde{w}_{i,i}^v)^T \widetilde{w}_{i,i}^v \mathbf{q}_i + \beta_3 \sum_{i=1}^{N} \sum_{j=1}^{N} \mathbf{q}_i^T \mathbf{q}_i z_{i,j} + \beta_3 \sum_{i=1}^{N} \mathbf{q}_i^T \sum_{j=1}^{N} \mathbf{q}_j z_{i,j} + \beta_5 \sum_{i=1}^{N} \sum_{j=1}^{N} \mathbf{q}_i^T \mathbf{q}_i s_{i,j}^v \right)^{-1} \left( \sum_{i=1}^{N} a^v \mathbf{q}_i^T (\widetilde{w}_{i,i}^v)^T \widetilde{w}_{i,i}^v \tilde{\mathbf{y}}_i + \beta_5 \mathbf{q}_i^T \sum_{j=1}^{N} \mathbf{b}_j s_{i,j}^v \right) \quad (18)$$

$$\mathbf{q}_i = (\tilde{\mathbf{x}}_{g,i}^v + \tilde{e}_{i,i}^v \tilde{\mathbf{h}}_i^v) \quad (19)$$

$$\mathbf{b}_i = \sum_{t=1, t \neq v}^{V} a^t (\tilde{\mathbf{x}}_{g,j}^t + \tilde{e}_{j,j}^t \tilde{\mathbf{h}}_j^t) \mathbf{P}_g^t \quad (20)$$

3) Update $\tilde{\mathbf{h}}_i^v$ with $\mathbf{P}_g^v$, $\mathbf{y}_i^u$ and $a^v$ fixed

Keeping $\mathbf{P}_g^v$, $\mathbf{y}_i^u$ and $a^v$ fixed, the following objective function should be minimized:

$$\min_{\tilde{\mathbf{h}}_i^v} \sum_{v=1}^{V} \sum_{i=1}^{N} a^v \left\| \widetilde{w}_{i,i}^v \left( (\tilde{\mathbf{x}}_{g,i}^v + \tilde{e}_{i,i}^v \tilde{\mathbf{h}}_i^v) \mathbf{P}_g^v - \tilde{\mathbf{y}}_i \right) \right\|^2 +$$
$$\beta_3 \sum_{v=1}^{V} \sum_{i=1}^{N} \sum_{j=1}^{N} z_{i,j} \left\| (\tilde{\mathbf{x}}_{g,i}^v + \tilde{e}_{i,i}^v \tilde{\mathbf{h}}_i^v) \mathbf{P}_g^v - (\tilde{\mathbf{x}}_{g,j}^v + \tilde{e}_{j,j}^v \tilde{\mathbf{h}}_j^v) \mathbf{P}_g^v \right\| + \beta_5 \sum_{v=1}^{V} \sum_{i=1}^{N} \sum_{j=1}^{N} s_{i,j}^v \left\| (\tilde{\mathbf{x}}_{g,i}^v + \tilde{e}_{i,i}^v \tilde{\mathbf{h}}_i^v) \mathbf{P}_g^v - \sum_{t \neq v}^{V} a^t (\tilde{\mathbf{x}}_{g,j}^t + \tilde{e}_{j,j}^t \tilde{\mathbf{h}}_j^t) \mathbf{P}_g^t \right\| \quad (21)$$

Taking the derivative of (21) with respect to $\tilde{\mathbf{h}}_i^v$ and setting it to zero, the update rule for $\tilde{\mathbf{h}}_i^v$ can be obtained as follows:

$$\tilde{\mathbf{h}}_i^v = \left( (a^v \widetilde{w}_{i,i}^{v\,T} \widetilde{w}_{i,i}^v + \sum_{j=1}^{N} (\beta_3 z_{i,j} + \beta_5 s_{i,j}^v)) \tilde{e}_{i,i}^{v\,T} \tilde{e}_{i,i}^v \right)^{-1} \left( a^v (\tilde{e}_{i,i}^v \widetilde{w}_{i,i}^v)^T \widetilde{w}_{i,i}^v \tilde{\mathbf{y}}_i (\mathbf{P}_g^v)^T + \beta_3 \sum_{j=1}^{N} \tilde{\mathbf{h}}_j^v s_{i,j}^v \mathbf{P}_g^v \mathbf{P}_g^{v\,T} + \beta_5 \sum_{j=1}^{N} \mathbf{b}_j s_{i,j}^v \mathbf{P}_g^{v\,T} \right) \left( \mathbf{P}_g^v \mathbf{P}_g^{v\,T} \right)^{-1} \quad (22)$$

$$\mathbf{b}_j = \sum_{t=1, t \neq v}^{V} a^t (\tilde{\mathbf{x}}_{g,j}^t + \tilde{e}_{j,j}^t \tilde{\mathbf{h}}_j^t) \mathbf{P}_g^t \qquad (23)$$

4) Update $\mathbf{y}_i^u$ with $\mathbf{P}_g^v$, $a^v$ and $\tilde{\mathbf{h}}_i^v$ fixed

Keeping $\mathbf{P}_g^v$, $\tilde{\mathbf{h}}_i^v$ and $a^v$ fixed, the following objective function should be minimized:

$$\min_{\mathbf{y}_i^u} \sum_{v=1}^{V} \sum_{i=1}^{N_u} a^v \left\| w_{i,i}^{u,v} \left( (\mathbf{x}_{g,i}^{u,v} + e_{i,i}^{u,v} \mathbf{h}_i^{u,v}) \mathbf{P}_g^{u,v} - \mathbf{y}_i^u \right) \right\|^2 + \beta_4 \sum_{v=1}^{V} \sum_{i=1}^{N_u} \sum_{j=1}^{N_u} \|\mathbf{y}_i^u - \mathbf{y}_j^u\| s_{i,j}^{u,v}$$

$$s.t. \mathbf{y}_i^u \geq 0, \ \sum_i^c \mathbf{y}_i^u = 1 \qquad (24)$$

Taking the derivative of (24) with respect to $\mathbf{y}_i^u$ and setting it to zero, the update rule for $\mathbf{y}_i^u$ can be obtained as follows:

$$\mathbf{y}_i^u = \theta^{-1}(\mathbf{f} + (\theta^{-1})^{-1}(1 - \sum_{i=1}^{c} \mathbf{f}^{-1})\mathbf{1}) \qquad (25)$$

$$\theta = \sum_{v}^{V} \sum_{i=1}^{N_u} a^v w_{i,i}^{u,v} w_{i,i}^{u,v} + \beta_4 \sum_{v}^{V} \sum_{i=1}^{N_u} \sum_{j=1}^{N_u} s_{i,j}^{u,v} \qquad (26)$$

$$\mathbf{f} = \sum_{v}^{V} \sum_{i=1}^{N_u} a^v w_{i,i}^{u,v} w_{i,i}^{u,v} (\mathbf{x}_{g,i}^{u,v} + e_{i,i}^{u,v} \mathbf{h}_i^{u,v}) \mathbf{P}_g^v + \beta_4 \sum_{j=1}^{N_u} \mathbf{y}_j^u s_{i,j}^{u,v} \qquad (27)$$

where $\mathbf{1} \in R^{1 \times c}$ is the all-one vector.

***Remark1:*** It should be noted that in model training, the error matrix $\{\tilde{\mathbf{H}}^v\}$ and pseudo label $\mathbf{Y}^u$ will be updated in each iteration, and the corresponding similarity matrix $\{\mathbf{S}^v\}$ and $\mathbf{Z}$ also need to be updated in each iteration. Besides, for the start of iterations, the pseudo label and error matrix are initialized with random value.

By optimizing (16), (18), (22) and (25) iteratively, the local optimal solution can be obtained, and the algorithm of the proposed SSIMV_TSK is given in Algorithm 1.

---

**Algorithm 1: SSIMV_TSK**

**Input:** incomplete multi-view dataset $\{\mathbf{X}^v\}$, $v=1,2…V$; the number of fuzzy rules $K$; number of maximum iterations $T$, the regularization parameters $\beta_1, \beta_2, \beta_3, \beta_4$ and $\beta_5$.
**Output:** $\mathbf{P}_g^v$, $a^v$ and $\mathbf{Y}^U$.
1: Use the Var-Part clustering algorithm to calculate the antecedent parameters of the TSK fuzzy systems for different views.
2: Use (3.b) - (3.d) to construct a new dataset $D_l^v = \{\mathbf{X}_g^{l,v}, \mathbf{Y}^l\}$ and $D_u^v = \{\mathbf{X}_g^{u,v}\}$ in the new feature space generated by fuzzy rules for each view. Let $\tilde{\mathbf{X}}_g^v = [\mathbf{X}_g^{l,v}; \mathbf{X}_g^{u,v}]$.
3: Initialize $\mathbf{P}_g^v$, $a^v$, $\mathbf{Y}^u$, $\tilde{\mathbf{H}}^v$. Let $\tilde{\mathbf{Y}} = [\mathbf{Y}^l; \mathbf{Y}^u]$.
4: for $t=1, 2, …, T$ do
5:   for $v=1, 2, …, V$ do
6:     Update $a^v$ based on (16).
7:     Update $\mathbf{P}_g^v$ based on (18).
8:     for $i=1, 2, …, N$ do
9:       Update $\tilde{\mathbf{h}}_i^v$ based on (22).
10:     end
11:   end
12:   for $i=1, 2, …, N_u$ do
13:     Update $\mathbf{y}_i^u$ based on (25).
14:   end
15: end

---

### G. Complexity Analysis

The complexity analysis of the SSIMV_TSK is discussed in this subsection. The time complexity of step 1 is $O(2Nd^vK)$, where $K$ is the number of fuzzy rules, $N$ is the number of instances, and $d^v$ is the number of features of the $v$-th view. The complexity of step 2 is $O(NK(d^v + 1))$. The time complexity of updating $a^v$ in step 6 is $O((Nd_g^v + d_g^vC + NC)NVT)$, where $d_g^v$ is the number of features of the $v$-th view in the new feature space generated by fuzzy rules, and $C$ is the number of classes. The time complexity of updating $\mathbf{P}_g^v$ in step 7 is $O((CN + d_g^vC + d_g^v + 1)Nd_g^vVT)$. The time complexity of updating $\tilde{\mathbf{h}}_i^v$ and $\mathbf{y}_i^u$ in step 9, 13 are $O((1 + d_g^v + d_g^v(1 + C) + CN)d_g^vNVT)$ and $O((d_g^vC + 1)N_uVT)$, respectively. $N_u$ is the number of unlabeled data. Since $N \gg C, N \gg d_g^v$, hence, the overall computational cost of Algorithm 1 is $O(N^2VT)$.

Table I Statistics of datasets

| Dataset | Size | Number of Views (Dimensions) | Number of Classes |
|---|---|---|---|
| Dermatology | 366 | 2(22-12) | 6 |
| Forest Type | 523 | 2(18-9) | 4 |
| Epileptic EEG | 500 | 2(20-6) | 2 |
| Multiple Features | 2000 | 2(76-47) | 10 |
| NUS-WIDE | 1000 | 2(144-73) | 5 |
| Caltech7 | 1474 | 3(48-40-254) | 7 |
| MSRCv1 | 210 | 3(24-256-254) | 7 |
| Image Segmentation | 2310 | 2(10-9) | 7 |

## IV. EXPERIMENTAL STUDIES

### A. Datasets

To evaluate the performance of the proposed method, eight benchmarking multi-view datasets were used in our experiments, of which four datasets were obtained from the UCI repository[1], and the remaining are obtained from the Epileptic EEG dataset [11], NUS-WIDE dataset [34], MSRCv1 dataset [35] and Caltech7 dataset [36], respectively. A brief description of these datasets is given below and in Table I.

1) Dermatology: It is a multi-view erythemato-squamous diseases dataset containing histopathological view and clinical view.
2) Forest Type: It is a multi-view forest remote sensing images dataset [37] containing image band view and spectral view.
3) Epileptic EEG: It is the Epileptic EEG dataset where discrete wavelet transform features and wavelet packet decomposition features were extracted as two views.
4) Multiple Features: It is a handwritten numerals dataset

---

[1] https://archive.ics.uci.edu/ml

where Karhunen-love values feature and Zernike moment feature were extracted as two views.
5) NUS-WIDE: It is an image classification dataset where color histogram and edge orientation histogram were extracted as two views.
6) Caltech7: It is a sub-dataset of Caltech101 [36] where Gabor, CENTRIST and wavelet texture features were extracted as three views.
7) MSRCv1: It is a scene recognition dataset where CMT [38], LBP [39], and GENT [40] features were extracted as three views.
8) Image Segmentation: It is a multi-view outdoor images dataset containing shape view and RBG view.

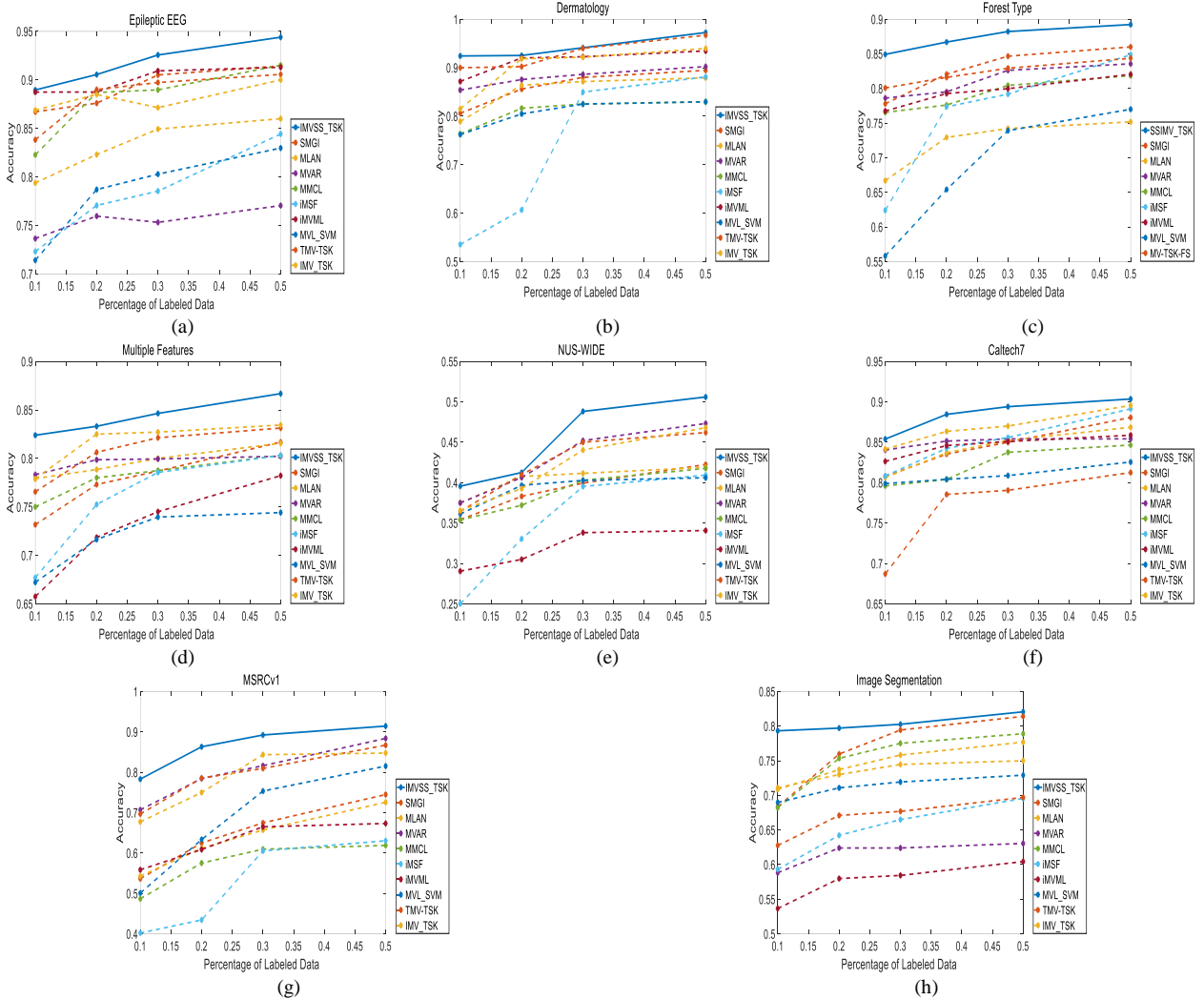

Fig. 2. The classification results of all algorithms under different rate of labeled data.

### B. Experimental Settings

The effectiveness of the proposed SSIMV_TSK was evaluated by comparing it with nine algorithms. Among the nine algorithms, four algorithms, i.e., IMSF [41], MVL-SVM [42], iMVML [43], IMV_TSK [26] are incomplete multi-view classification algorithms, and the rest five algorithms, i.e., SMGI [15], MLAN [16], MVAR [17], MMCL [44] and TMV-TSK [19] are semi-supervised multi-view classification algorithms. The brief descriptions of the nine methods are given below:

1) IMSF: This algorithm transforms the incomplete multi-view problem into a multi-task problem by dividing the incomplete multi-view dataset into several groups of data.
2) MVL-SVM: This algorithm randomly selects landmark for similarity learning and missing view imputation, and then uses multi-view SVM for classification.
3) iMVML: This algorithm learns a common view between views, and then models the hidden view.
4) IMV_TSK: This algorithm integrates the common representation learning and missing view imputation into one process and then uses them to construct a

- multi-view model.
5) SMGI: Based on label propagation theory [45], this algorithm carries out semi-supervised learning and sparse learning on each view and then unifies the results of each view to achieve the final results.
6) MALN: This algorithm integrates the label propagation learning and common view learning into one process, and introduces the adaptive view weight to improve the robustness of the model.
   MVAR: This algorithm constructs a global regression model for each view and gets the final decision by the weighted combination of the results of multiple views.
7) MMCL: This algorithm integrates the label propagation learning and course learning [46], and takes the pseudo label of each round as feedback to guide the model learning.
8) TMV-TSK: This algorithm learns both the model and the pseudo label simultaneously, and introduces the matrix factorization to further improve the performance of the model.

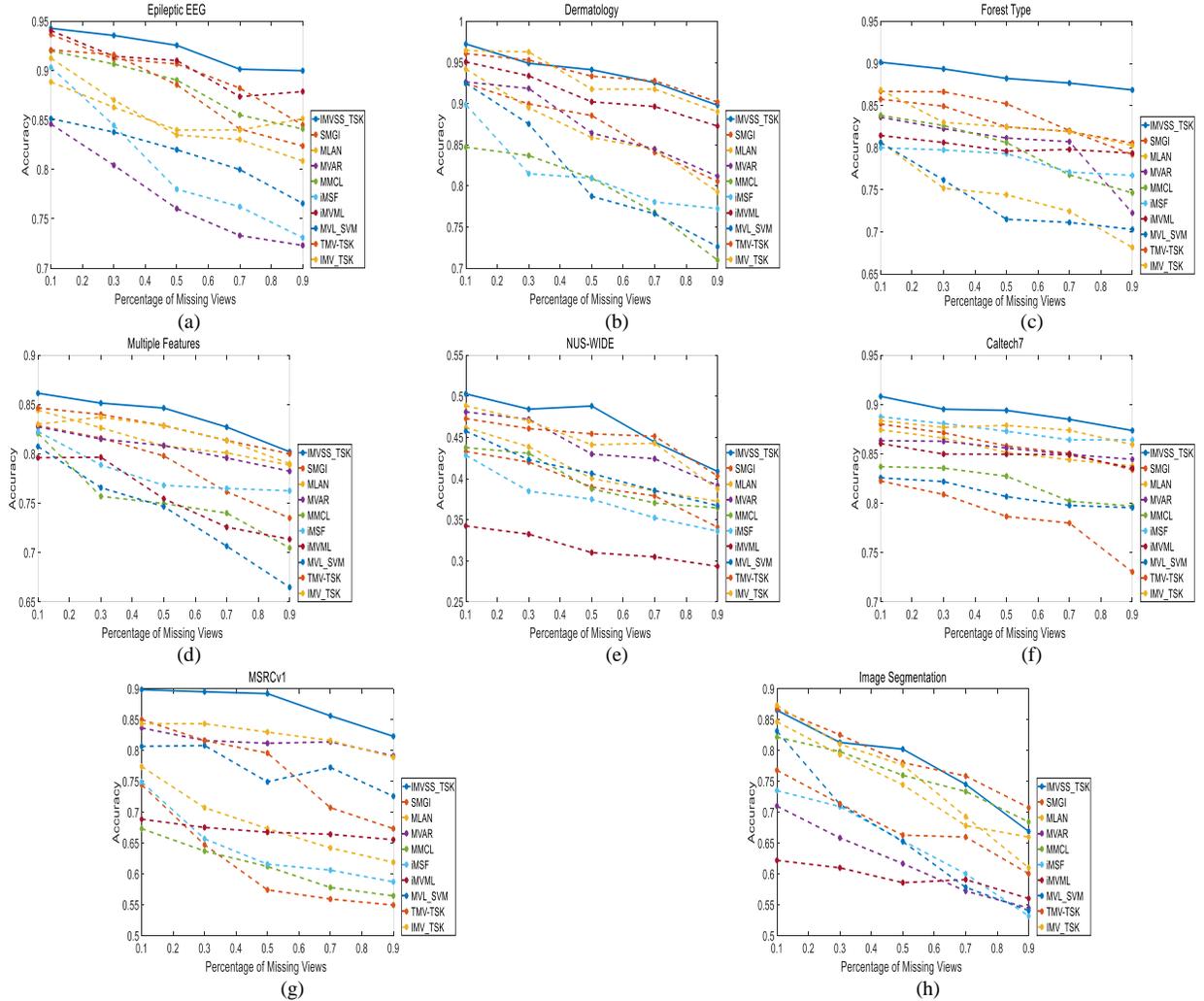

Fig. 3. The classification results of all algorithms under different rate of missing views.

For fair comparison, all regularization parameters were set within $\{2^{-5}, 2^{-4},…,2^4,2^5\}$. The number of the fuzzy rule was set between 2 and 10 with a step size of 2, and the number of nearest neighbors was set to 7. In order to compare the proposed SSIMV-TSK with adopted nine algorithms effectively, we conducted two types of different experiments, i.e., the semi-supervised multi-view classification and the incomplete multi-view classification.

(1) For the semi-supervised multi-view classification, 10%-50% data were randomly chosen as labeled data and the rest were used as unlabeled data. Meanwhile, the rate of missing views was fixed at 50%.

(2) For the incomplete multi-view classification, 10%-90% instances of each view were randomly removed, with a step

size of 20%. Meanwhile, the labeled data was fixed at 30% and the rest was used as unlabeled data.

To be fair, we evaluate the performance on unlabeled data rather than unseen data, as with most transductive methods. The imputation method [20] is applied to five semi-supervised multi-view classification algorithms, since they cannot deal with incomplete multi-view data directly. To reduce the potential bias due to the randomly generated dataset, the above process was repeated 20 times and the average accuracy was taken as the final result for comparison.

## C. Experimental Results

### 1) Results of Semi-Supervised Classification under the Different Rate of Labeled Data

The classification results of all algorithms under the different rate of labeled data are shown in Fig. 2.

From Fig. 2, we can see that SSIMV_TSK has a great advantage over other algorithms in most cases, which indicates that the bidirectional structural preservation of instance and label and adaptive multiple alignment collaborative learning used in this paper address the few labeled incomplete multi-view data problem effectively. Meanwhile, even missing views are imputed, the performance of the traditional semi-supervised multi-view algorithms are still poor in most cases, which indicates that it is not an effective way to separate the missing view imputation from the modeling process.

### 2) Results of Incomplete Multi-view Classification under the Different Missing Rate of Views

Fig. 3 shows the classification results of all algorithms under the different missing rate of views.

It can be seen from Fig. 3 that SSIMV_TSK is superior to other incomplete multi-view algorithms in most cases. Meanwhile, we can see that when there are few missing views, SSIMV_TSK has a smaller advantage over other algorithms. However, the advantage of SSIMV_TSK increases with the increase of missing views. These results indicate that integrating the missing view imputation and the model construction into one process, as well as constructing the bidirectional structural preservation of instance and label and adaptive multiple alignment collaborative learning can effectively improve the robustness of the model, especially when the percentage of missing view is large.

Table II Friedman test results.

| Algorithm | Ranking | $p$-value | Hypothesis |
|---|---|---|---|
| SSIMV_TSK | 1 | | |
| IMV_TSK | 3.375 | | |
| TMV-TSK | 4.25 | | |
| SMGI | 4.875 | | |
| MVAR | 5.75 | 0.00001 | Reject |
| iMVML | 6.125 | | |
| MLAN | 6.125 | | |
| iMSF | 7.375 | | |
| MMCL | 7.375 | | |
| MVL-SVM | 8.75 | | |

Table III Post-holm results (reject hypothesis if $p$-value <0.025)

| $i$ | Algorithm | $Z = (\mathbf{R}_o - \mathbf{R}_i)/SE$ | $p$ | Holm = $\alpha/i$, $\alpha = 0.05$ | Null Hypothesis |
|---|---|---|---|---|---|
| 9 | MVL-SVM/ SSIMV_TSK | 5.119482 | 0 | 0.005556 | Reject |
| 8 | MMCL/ SSIMV_TSK | 4.211186 | 0.000025 | 0.00625 | Reject |
| 7 | iMSF/ SSIMV_TSK | 4.211186 | 0.000025 | 0.007143 | Reject |
| 6 | MLAN/ SSIMV_TSK | 3.85464 | 0.000711 | 0.008333 | Reject |
| 5 | iMVML / SSIMV_TSK | 3.85464 | 0.000711 | 0.01 | Reject |
| 4 | MVAR / SSIMV_TSK | 3.237747 | 0.001703 | 0.0125 | Reject |
| 3 | SMGI/ SSIMV_TSK | 2.559741 | 0.010475 | 0.016667 | Reject |
| 2 | TMV-TSK/ SSIMV_TSK | 2.146879 | 0.031803 | 0.025 | Not Reject |
| 1 | IMV_TSK/ SSIMV_TSK | 1.568873 | 0.116677 | 0.05 | Not Reject |

## D. Statistical Analysis

In this subsection, the Friedman test and the post-hoc Holm test were performed to analyze the effectiveness of all algorithms when the rate of missing views was 50% and the labeled data was 30%.

To evaluate whether the performance differences between the ten algorithms on all datasets were statistically significant, the Friedman test [47] was first performed. The null hypothesis of Friedman tests is defined as no difference in performance between all algorithms. When the $p$-value of the test is less than 0.05, the difference in classification performance between the ten algorithms is statistically significant. Otherwise, there is no statistically significant difference in classification performance. As shown in Table II, the $p$-value is far less than 0.05, so the difference between algorithms is statistically significant. Moreover, the ranking of SSIMV_TSK is superior to other algorithms, which means SSIMV_TSK is the best among the ten algorithms.

Then, we performed the post-hoc Holm test to further compare SSIMV_TSK with the other nine algorithms. If the $p$-value less than the Holm value, the difference in classification performance between the two algorithms is statistically significant. Otherwise, there is no statistically significant difference. From Table III, it can be seen that SSIMV_TSK has a statistically significant difference with most algorithms.

Although the null hypothesis is not rejected when compared with TMV-TSK and IMV_TSK, the results in Figs. 2 and 3 show that IMV_TSK still outperforms TMV-TSK and IMV_TSK.

*E. Effectiveness Analysis*

In this subsection, the effectiveness of the bidirectional structural preservation of instances and labels, as well as the adaptive multiple alignment collaborative learning are analyzed when the rate of missing views is 50% and the labeled data is 30%. We denote SSIMV_TSK without the adaptive multiple alignment collaborative learning as SSIMV_TSK1 and denote SSIMV_TSK without the bidirectional structural preservation of instances and labels as SSIMV_TSK2.

It can be clearly seen that SSIMV_TSK outperforms SSIMV_TSK1 and SSIMV_TSK2 for most datasets, indicating the effectiveness of the bidirectional structural preservation of instances and labels, as well as the adaptive multiple alignment collaborative learning in SSIMV_TSK. Besides, SSIMV_TSK2 outperforms SSIMV_TSK1 for most datasets, which indicates that using the bidirectional structural preservation of instances and labels is more effective to improve the robustness of the model under the incomplete and few labeled data multi-view scenario.

Table IV Accuracy of SSIMV_TSK1, SSIMV_TSK2 and SSIMV_TSK

| Datasets | SSIMV_TSK1 | SSIMV_TSK2 | SSIMV_TSK |
|---|---|---|---|
| Dermatology | 0.8941±0.0305 | 0.9098±0.0221 | **0.9412±0.0139** |
| Forest Type | 0.8384±0.0214 | 0.8521±0.0039 | **0.8822±0.0625** |
| Epileptic EEG | 0.9114±0.0141 | 0.92±0.0101 | **0.9257±0.0099** |
| NUS-WIDE | 0.4714±0.0273 | 0.4743±0.0012 | **0.4881±0.0223** |
| Multiple Features | 0.8207±0.0131 | 0.8121±0.0116 | **0.8464±0.0016** |
| Caltech7 | 0.8774±0.0069 | 0.8865±0.0041 | **0.8940±0.0034** |
| MSRCv1 | 0.8798±0.0337 | 0.8653±0.0288 | **0.8903±0.0150** |
| Image Segmentation | 0.7879±0.0166 | 0.7854±0.0144 | **0.8023±0.0241** |
| Average | 0.8101±0.0205 | 0.8140±0.0120 | **0.8338±0.0191** |

*F. Interpretability Analysis*

In this section, we take the results of SSIMV_TSK on the Epileptic EEG dataset with 50% missing views and 30% labeled data as an example to analyze the interpretability of SSIMV_TSK. In the experiment, the rule number was set to four, *i.e.*, *K*=4. Table S1 and Fig. S1 of the *Supplementary Materials* section show the four rules for the WPD view of the trained SSIMV_TSK and the corresponding four membership functions, respectively. According to the order of the membership center, the linguistic descriptions of these four membership functions can be expressed as *High*, *Little High*, *Medium* and *Low*. These linguistic expressions are the IF-part, and the corresponding linear function is THEN-part. Combing them, four fuzzy rules can be defined. For example, the first fuzzy rule of the WPD view can be expressed as follows:

*IF the energy of the EEG signal in the frequency Band 1 is Little High, and the energy of the EEG signal in the frequency Band 2 is High, and the energy of the EEG signal in the frequency Band 3 is High, and the energy of the EEG signal in the frequency Band 4 is Little High, and the energy of the EEG signal in the frequency Band 5 is Medium, and the energy of the EEG signal in the frequency Band 6 is Medium, THEN the decision values of the four outputs are given as follows:*

$$\mathbf{f}^1(\mathbf{x}) = \begin{bmatrix} 0.0993+0.2151x_1-0.4470x_2-0.5105x_3 \\ -0.0583x_4+1.701x_5+0.2224x_6, \\ 0.7563+0.1070x_1+0.5184x_2+0.3370x_3 \\ -0.0619x_4-0.8593x_5-0.0419x_6 \end{bmatrix}$$

The explanation of the DWT view can be analyzed in the same way. Moreover, the use of the four rules generated by SSIMV_TSK are further explained in Fig. S2 of the *Supplementary Materials*.

V. CONCLUSION

To address the incomplete, few labeled multi-view data learning problem, a new semi-supervised incomplete multi-view TSK fuzzy system modeling method called SSIMV_TSK is proposed in this paper. Different from existing methods, SSIMV_TSK integrates the missing view imputation, pseudo label learning and model constructing into a single framework, which makes each task benefit from the others effectively. Meanwhile, a bidirectional structural preservation mechanism of instances and labels is proposed to improve the quality of the imputed missing view and the pseudo label. Besides, to mine the consistency and complementarity information between the incomplete, few labeled multi-view data efficiently, a new adaptive multiple alignment collaborative learning is also proposed. Moreover, the proposed model is built based on fuzzy logic rules, which alleviates the problem of poor interpretability in many exiting methods. The results of extensive experiments also verify the advantages of the proposed method.

Although the proposed SSIMV_TSK method has demonstrated promising performance, there is still room for further study. For example, there are many regularization parameters in current models, which leads to a long time for determining the optimal parameters. We will study how to reduce the time for choosing the optimal parameters and guarantee the performance of the model simultaneously. Moreover, constructing the similarity matrix is time-consuming step of the proposed method, how to reduce the time for the similarity matrix construction will be very valuable. One more thing deserves to note is that although the proposed SSIMV_TSK is a transductive semi-supervised method, it can be applied in some inductive tasks. For example, when the test data are complete multi-view data, the classification results can be predicted directly using the trained model. However, when the test data are incomplete multi-view data, the trained model is unavailable. How to make the proposed method applicable to this situation will be addressed in our future work.